\documentclass{article}

\usepackage{microtype}
\usepackage{graphicx}
\usepackage{subfigure}
\usepackage{booktabs} %

\usepackage[hypertexnames=false]{hyperref}

\usepackage[accepted]{icml2023}

\usepackage{amsmath}
\usepackage{amssymb}
\usepackage{mathtools}
\usepackage{amsthm}

\usepackage[capitalize,noabbrev]{cleveref}

\usepackage{enumitem}
\usepackage{multirow}
\usepackage[htt]{hyphenat}
\usepackage{xcolor}
\definecolor{mygreen}{HTML}{257D00}

\theoremstyle{plain}

\theoremstyle{definition}

\theoremstyle{remark}

\usepackage[textsize=tiny]{todonotes}

\icmltitlerunning{Evaluating GPT-3.5 and GPT-4 on Grammatical Error Correction for Brazilian Portuguese}

\begin{document}

\twocolumn[
\icmltitle{Evaluating GPT-3.5 and GPT-4 on \\ Grammatical Error Correction for Brazilian Portuguese}

\icmlsetsymbol{equal}{*}

\begin{icmlauthorlist}
\icmlauthor{Maria Carolina Penteado}{equal}
\icmlauthor{Fábio Perez}{equal}
\end{icmlauthorlist}

\icmlcorrespondingauthor{Fábio Perez}{fabiovmp@gmail.com}

\icmlkeywords{Machine Learning, ICML}

\vskip 0.3in
]

\printAffiliationsAndNotice{\icmlEqualContribution} %

\begin{abstract}
We investigate the effectiveness of GPT-3.5 and GPT-4, two large language models, as Grammatical Error Correction (GEC) tools for Brazilian Portuguese and compare their performance against Microsoft Word and Google Docs. We introduce a GEC dataset for Brazilian Portuguese with four categories: Grammar, Spelling, Internet, and Fast typing. Our results show that while GPT-4 has higher recall than other methods, LLMs tend to have lower precision, leading to overcorrection. This study demonstrates the potential of LLMs as practical GEC tools for Brazilian Portuguese and encourages further exploration of LLMs for non-English languages and other educational settings.
\end{abstract}

\section{Introduction}

Large language models (LLMs) have revolutionized the field of natural language processing by enabling computers to process and generate human-like language~\cite{kasneci2023chatgpt}. LLMs have the potential to be particularly useful for Grammatical Error Correction (GEC)~\cite{wu2023chatgpt_or_grammarly, bryant2022grammatical_survey} and can be a valuable educational tool to enhance students' writing skills by providing real-time feedback and corrections.

Traditional GEC methods usually rely on pre-defined rules to identify and correct errors. While these methods can effectively detect simple misspellings, they may struggle to correct more complex grammatical errors. In contrast, LLMs can model language from large amounts of text data, which could lead to more natural and contextually appropriate corrections. By analyzing the context and meaning of a sentence, LLMs may identify errors that traditional methods may miss and provide more nuanced corrections.

Although large language models (LLMs) have gained widespread attention for their performance in English language applications, recent studies have shown that they can produce good results for other languages. While the amount of data available for training LLMs in languages other than English is often more limited, the success of these models in tasks such as translation, language modeling, and sentiment analysis demonstrates their potential for improving language processing across a range of different languages.

In this work, we take an initial step on investigating the effectiveness of GPT-3.5 and GPT-4~\cite{openai2023gpt4}, two LLMs created by OpenAI, as a GEC tool for Brazilian Portuguese. Our main contributions are the following:

\begin{enumerate}[leftmargin=*]
    \itemsep0em
    \item We compare GPT-3.5 and GPT-4 against Microsoft Word and Google Docs and show that LLMs can be a powerful tool for GEC.%
    \vspace{-0.1cm}
    \item We crafted a GEC dataset for Brazilian Portuguese, including four categories:  \textit{Grammar}, \textit{Spelling}, \textit{Internet}, and \textit{Fast typing}. \vspace{-0.1cm}
    \item We quantitative and qualitatively evaluated LLMs as a GEC tool for Brazilian Portuguese. %
\end{enumerate}

\section{Related Work}

\citet{nunes2023evaluating_enem} explored the use of GPT-3.5 and GPT-4 to answer questions for the \textit{Exame Nacional do Ensino Médio} (ENEM), an entrance examination used by many Brazilian universities. They tested different prompt strategies, including using Chain-of-Thought (CoT) to generate explanations for answers, and found that GPT-4 with CoT was the best-performing approach, achieving an accuracy of 87\% on the 2022 exam.

\citet{wu2023chatgpt_or_grammarly} evaluated the performance of different models for GEC, including Grammarly, GECToR, and ChatGPT (authors did not specify whether they used GPT-3.5 or GPT-4), and found that automatic evaluation methods result in worse numbers for ChatGPT than other GEC methods. In contrast, human evaluation showed that ChatGPT produces fewer under-corrections or miscorrections and more overcorrections, indicating not only the potential of LLMs for GEC but also the limitation of automatic metrics to evaluate GEC tools.

\citet{fang2023chatgpt_highly_fluent_gec} investigated GPT-3.5's potential for GEC using zero- and few-shot chain-of-thought settings. The model was evaluated in English, German, and Chinese, showcasing its multilingual capabilities. The study found that GPT-3.5 exhibited strong error detection and generated fluent sentences but led to over-corrections.

Despite their outstanding performance on many tasks, LLMs may not be the silver bullet for NLP in multi-lingual settings. \citet{lai2023chatgpt_beyond_english} evaluated ChatGPT on various NLP tasks and languages, showing that it performs significantly worse than state-of-the-art supervised models for most tasks in different languages, including English. Their work does not include GEC, and Portuguese is only evaluated for relation extraction.

The shortage of academic research on LLMs for multi-lingual settings, especially for Brazilian Portuguese, highlights the need for further engagement in this field. Our work aims to fill this gap by exploring the potential of GPT-3.5 and GPT-4 as GEC tools for Brazilian Portuguese.

\section{Methodology}

\subsection{Dataset}

We created the dataset (Table~\ref{table:dataset}) by having native Brazilian Portuguese speakers manually write multiple sentences and dividing them into four categories: grammar, orthography, mistypes, and internet language. All categories list incorrect sentences and their correct pairs. The categories are described as follows:

\begin{itemize}[leftmargin=*]
    \itemsep0em 
    \item \textbf{Grammar} --- 34 sets of three (totaling 102) phrases containing two words or expressions that are commonly swapped due to their similarity. \vspace{-0.10cm}
    \item \textbf{Spelling} --- 100 sentences with spelling, punctuation, or accentuation errors. \vspace{-0.10cm}
    \item \textbf{Fast typing} --- 40 mistyped (e.g., when typing too fast) sentences. \vspace{-0.10cm}
    \item \textbf{Internet language} --- 40 sentences containing slang, abbreviations, and neologisms often used in virtual communication. %
\end{itemize}

We find it important to acknowledge that the dataset may reflect the biases of the human curators and may not fully encompass the complexities and variations present in real-world data. However, the limited availability of corpora specifically designed for GEC in Brazilian Portuguese compelled us to create our dataset, which, despite its potential limitations, represents a starting point in the task.

The dataset is available in the supplementary material.

\begin{table*}[t]
\caption{Description of the developed dataset, divided into four categories: \textit{Grammar}, \textit{Spelling}, \textit{Fast typing}, and \textit{Internet}. The table shows the total correct and incorrect phrases per category and example phrases from the dataset.
\textit{Grammar} and \textit{Spelling} include only one error per phrase, while \textit{Fast typing} and \textit{Internet} include multiple.}
\label{table:dataset}
\begin{center}
\begin{small}
\begin{sc}
\begin{tabular}{lrrp{0.34\linewidth}p{0.34\linewidth}}
\toprule
& \textbf{Cor.} & \textbf{Inc.} & \textbf{Corr. Example} & \textbf{Incorr. Example} \\ \midrule
Grammar& 102 & 102 & \begin{tt}Você nunca mais falou com \textcolor{red}{agente}\end{tt} & \begin{tt}Você nunca mais falou com \textcolor{mygreen}{a gente}\end{tt} \\ \midrule
Spelling & 100 & 100 & \begin{tt}A \textcolor{red}{análize} do documento será feita por um advogado\end{tt} & \begin{tt}A \textcolor{mygreen}{análise} do documento será feita por um advogado\end{tt} \\ \midrule
Fast typing & - & 40 & \begin{tt}ele já \textcolor{red}{quebru} todos \textcolor{red}{oc} copos \textcolor{red}{nvoos} que \textcolor{red}{comepri} esse \textcolor{red}{mêsd}\end{tt} & \begin{tt}Ele já \textcolor{mygreen}{quebrou} todos \textcolor{mygreen}{os} copos \textcolor{mygreen}{novos} que \textcolor{mygreen}{comprei} esse \textcolor{mygreen}{mês}\end{tt} \\ \midrule
Internet & - & 40 & \begin{tt}\textcolor{red}{n} dá \textcolor{red}{p} escutar, \textcolor{red}{n} sei o \textcolor{red}{pq}\end{tt} & \begin{tt}\textcolor{mygreen}{Não} dá \textcolor{mygreen}{para} escutar, \textcolor{mygreen}{não} sei o \textcolor{mygreen}{porquê}\end{tt} \\ \bottomrule 
\end{tabular}
\end{sc}
\end{small}
\end{center}
\vskip -0.1in
\end{table*}

\subsection{Experiments}

\begin{table}[t]
\caption{Prompt used for GPT-3.5 and GPT-4 and its English translation as reference. We prompted both models to add \textit{[Correta]} if the phrase is correct to avoid them appending long texts saying the phrase is correct. We removed any \textit{[Correta]} occurrence before evaluating the models.}
\label{table:prompts}
\begin{center}
\begin{small}
\begin{sc}
\begin{tabular}{p{0.95\linewidth}}
\toprule
\textbf{Prompt} \\
\midrule
\begin{tt}
Corrija os erros gramaticais das seguintes frases em Português brasileiro. Não altere o significado das frases, apenas as corrija. Não altere frases gramaticalmente corretas, apenas escreva [Correta] após a frase.

\vspace{0.3cm}

\{list of phrases\}
\end{tt}
\vspace{0.2cm} \\
\midrule
\textbf{Prompt translation to English} \\ \midrule
\begin{tt}
Fix the grammatical errors in the following Brazilian Portuguese sentences. Do not change the meaning of the sentences, just fix them. Do not change grammatically correct sentences, just write [Correct] after the sentence.
\end{tt}
\\
\bottomrule
\end{tabular}
\end{sc}
\end{small}
\end{center}
\vskip -0.1in
\end{table}

We compared GPT-3.5 and GPT-4, two LLMs, against the spelling and grammar error correction features on Google Docs and Microsoft Word, two widely-used text editors.

For Google Docs (\href{https://docs.google.com/}{docs.google.com}), we first set the language on \textit{File} $\rightarrow$ \textit{Language} $\rightarrow$ \textit{Português (Brasil)}. Then we selected \textit{Tools} $\rightarrow$ \textit{Spelling and grammar} $\rightarrow$ \textit{Spelling and grammar check}. Finally, for every error, we clicked on \textit{Accept}.

We used the online version of Microsoft Word (\href{https://onedrive.live.com/}{onedrive.live.com}). First, we set the language on \textit{Set Proofing Language} $\rightarrow$ \textit{Current Document} $\rightarrow$ \textit{Portuguese (Brazil)}. Then, we opened the \textit{Corrections} tab and selected all errors under \textit{Spelling and Grammar}. For each error, we selected the first suggestion. We repeated the process until Word stopped finding errors.

For GPT-3.5 and GPT-4, we used ChatGPT (\href{https://chat.openai.com/}{chat.openai.com}) with the prompt shown in Table~\ref{table:prompts}. We shuffled the phrases and ensured the same pair of correct and incorrect phrases did not appear in the same prompt. Instead of running phrases individually, we ran 20 to 26 simultaneous phrases in one prompt, depending on the category. We used the ChatGPT interface and not the OpenAI API since we did not have access to the GPT-4 API at the time of the experiments. We did not focus on optimizing the prompt as our goal is to evaluate the usefulness of LLMs for GEC in Brazilian Portuguese without requiring deep LLMs knowledge. We believe more careful prompt engineering may improve the results.

\section{Results}

\begin{table*}[t]
\caption{Evaluation results for all experiments. Since the results are not deterministic, values for GPT-3.5 and GPT-4 represent the average and standard deviation for three runs.}
\label{table:results}
\vskip 0.15in
\begin{center}
\begin{small}
\begin{sc}
\begin{tabular}{llrrrr}
\toprule
 &  & \multicolumn{1}{r}{\textbf{MS Word}} & \multicolumn{1}{r}{\textbf{Google Docs}} & \multicolumn{1}{r}{\textbf{GPT-3.5}} & \multicolumn{1}{r}{\textbf{GPT-4}} \\ \midrule
\textbf{Internet} & Recall & 12.5\% & 5.0\% & 78.3±1.3\% & \textbf{89.3±1.3\%} \\ \midrule
\textbf{Fast typing} & Recall & 27.5\% & 40.0\% & 85.0±0.0\% & \textbf{90.0±1.3\%} \\ \midrule
\textbf{Grammar} & Precision & \multicolumn{1}{r}{89.1\%} & \multicolumn{1}{r}{\textbf{97.4\%}} & \multicolumn{1}{r}{67.5±0.2\%} & \multicolumn{1}{r}{86.8±0.7\%} \\
 & Recall & 40.2\% & 36.3\% & 63.7±1.7\% & \textbf{75.5±1.7\%} \\
 & $F_{0.5}$ & 71.7\% & 72.8\% & 66.7±0.5\% & \textbf{84.3±1\%} \\
 & TNR & 95.1\% & \textbf{99.0\%} & 69.3±0.6\% & 88.5±0.6\% \\ \midrule
\multirow{4}{*}{\textbf{Spelling}} & Precision & \multicolumn{1}{r}{94.9\%} & \multicolumn{1}{r}{\textbf{100\%}} & \multicolumn{1}{r}{79.7±1.7\%} & \multicolumn{1}{r}{99.3±0.6\%} \\
 & Recall & 74.0\% & 66.0\% & 85±3.5\% & \textbf{92.0±6.1\%} \\
 & $F_{0.5}$ & 89.8\% & 90.7\% & 80.7±2\% & \textbf{97.7±1.8\%} \\
 & TNR & 96.0\% & \textbf{100\%} & 78.3±1.5\% & 99.3±0.6\% \\ \bottomrule
\end{tabular}
\end{sc}
\end{small}
\end{center}
\vskip -0.1in
\end{table*}

CoNLL2014~\cite{ng2014conll} employs an evaluation method in which GEC tools are evaluated by all edits they made on phrases against gold-standard edits. Instead, we evaluate GEC tools by comparing the modified phrases against the gold-standard ones. For the \textit{Grammar} and \textit{Spelling} categories, we also ran GEC tools on phrases without grammatical errors to evaluate false positives. We calculated four metrics:

\begin{itemize}[leftmargin=*]
    \itemsep0em 
    \item \textbf{Precision} --- From the phrases modified by the GEC tool, how many were successfully corrected? %
    \item \textbf{Recall} --- From the ungrammatical phrases, how many were successfully corrected by the GEC tool? %
    \item \textbf{$F_{0.5}$ Score} --- A metric that combines both precision and recall, but emphasizes precision twice as much as recall. It is commonly used in GEC studies~\cite{ng2014conll}. %
    \item \textbf{True Negative Rate (TNR)} --- From the grammatical phrases, how many were successfully not modified by the GEC tool? %
\end{itemize}

We evaluated \textit{Grammar} and \textit{Spelling} using the four metrics and \textit{Internet} and \textit{Fast typing} using recall. Table~\ref{table:results} shows the results for all experiments. We define true/false positive/negative as follows (see Table~\ref{table:tp_tn_fp_fn} for examples):

\vspace{-0.12cm}
\begin{itemize}[leftmargin=*]
    \itemsep0em 
    \item \textbf{True Positive (TP)} --- incorrect phrase is corrected by the GEC tool. \vspace{-0.10cm}
    \item \textbf{False Positive (FP)} --- correct phrase is wrongly modified by the GEC tool. \vspace{-0.10cm}
    \item \textbf{True Negative (TN)} --- correct phrase is not modified by the GEC tool. \vspace{-0.10cm}
    \item \textbf{False Negative (FN)} --- incorrect phrase is not corrected by the GEC tool. %
\end{itemize}

\section{Discussion}

Results (Table~\ref{table:results}) for \textit{Grammar} and \textit{Spelling} show that GPT-3.5 and GPT-4 have superior recall and worse precision than Microsoft Word and Google Docs. These results agree with those by \citet{wu2023chatgpt_or_grammarly} and \citet{fang2023chatgpt_highly_fluent_gec} and suggest that while GPT models are very effective at identifying errors, they tend to make more corrections than necessary, potentially altering the meaning or style of the text. The lower TNR values also confirms that LLMs tend to modify correct phrases.

One possible explanation for the higher recall of LLMs is their ability to model language from large amounts of text data, allowing them to capture a wide range of language patterns and contextual nuances. This makes them effective at detecting complex grammatical errors, but their open-ended nature can lead to overcorrection by generating multiple possible corrections without clearly picking the most appropriate one. Furthermore, LLMs may have lower precision because they often prioritize fluency and coherence over grammatical accuracy, leading to unnecessary changes to the text, increasing false positives. In contrast, rule-based methods prioritize grammatical accuracy and make changes only when necessary.

Although strongly impacted by the lower precision, GPT-4 shows a higher $F_{0.5}$ score than any other methods for both \textit{Grammar} and \textit{Spelling}. GPT-3.5, however, has lower $F_{0.5}$ scores than Google Docs and Microsoft Word, indicating that GPT-4 is a clear improvement over GPT-3.5 as a GEC tool for Brazilian Portuguese.

Finally, GPT-3.5 and GPT-4 perform much better than Microsoft Word and Google Docs for the \textit{Internet} and \textit{Fast typing} categories. Traditional methods struggle with these tasks as they are strongly context-dependent, while LLMs thrive due to being trained on vast amounts of text. This demonstrates the capabilities of LLMs as a GEC tool for non-traditional GEC scenarios.

We also performed a qualitative analysis by checking each correction provided by GPT-3.5 and GPT-4. We identified four explicit behaviors. See Table~\ref{table:behaviors} for examples of phrases for each behavior.

The first behavior (over-correction) considers extra edits that lead to grammatically correct sentences without meaning changes (e.g., add/remove commas, convert commas into semicolons, and upper vs. lower case). GPT-3.5 delivered 54 (out of 484) sentences with such behavior vs. six from GPT-4. The second behavior (omission) refers to models failing to detect errors and occurred 22 and 23 times on GPT-3.5 and GPT-4, respectively.

The third behavior (grammatical miscorrection) includes changes that adhere to grammatical rules but modify the sentence's meaning (e.g., removing/adding/substituting words and inverting the order of excerpts). GPT-3.5 corrections fell in this category 41 times vs. 13 times for GPT-4. Finally, the fourth behavior (ungrammatical miscorrection) is similar to the previous one but leads to ungrammatical sentences. GPT-3.5 and GPT-4 produced 3 and 1 outputs in this category, respectively.

\subsection{Limitations and Challenges of LLMs as GEC tools}

While large language models (LLMs) have shown considerable promise for Grammatical Error Correction (GEC), limitations and challenges must be considered when using these models for GEC tasks.

\begin{description}[wide=0\parindent]
   \item[Open-endedness] LLMs are open-ended and stochastic by nature. Unlike rule-based models, LLMs generate text based on patterns learned from training data. This can make it difficult to constrain the model, resulting in the replacement of grammatically correct words with other words that may occur more frequently in a given context~\cite{bryant2022grammatical_survey}. Another unpredictability of LLMs is their tendency to produce "hallucinations" -- outputs that are not necessarily true or based on the input data~\cite{openai2023gpt4}. This can result in the generation of incorrect or irrelevant corrections.
   \item[Prompt engineering] LLMs' performance rely on the used prompts~\cite{Brown2020LanguageModelsAreFewShotLearners}, where LLM-based GEC tools might need prompt engineering to achieve high-quality outputs. The effectiveness of a prompt may vary significantly depending on the task, and determining an optimal prompt may require extensive experimentation.
   \item[Hardware constraints] The large-scale nature of LLMs requires powerful hardware, which can be a barrier for many users and institutions. This can make LLMs less accessible and cost-effective for GEC tasks, particularly for those with limited resources or budget constraints. To interact with LLMs that cannot run on consumer hardware, one must send requests to third-party servers, requiring an internet connection and posing a privacy risk.
   \item[Biases and malicious uses] LLMs may contain biases and inaccuracies, posing a challenge in ensuring that corrections do not inadvertently perpetuate harmful stereotypes or misinformation~\cite{Blodgett2020LanguageIsPower, Nadeem2020StereoSetMS, GarridoMuoz2021ASO_Survey_Bias}. LLMs may also suffer from malicious attacks intent to misleading the model~\cite{perez2022ignore, Greshake2023MoreTY}.
\end{description}

\section{Conclusion}

Our study demonstrates the potential of LLMs as effective GEC tools for Brazilian Portuguese. We hope this work encourages further exploration of the impact of LLMs on Brazilian Portuguese and other non-English languages and spurs interest in developing and refining LLMs for diverse linguistic contexts. As a suggestion for future works, we believe that curating larger and better datasets that capture real-world data (e.g., by collecting grammatical errors made in real scenarios) could strengthen the field. Moreover, we encourage researchers to continue investigating the potential of LLMs in educational settings (see Appendix~\ref{sec:appendix_more_educational_applications}).

\bibliography{example_paper}

\begin{thebibliography}{14}
\providecommand{\natexlab}[1]{#1}
\providecommand{\url}[1]{\texttt{#1}}
\expandafter\ifx\csname urlstyle\endcsname\relax
  \providecommand{\doi}[1]{doi: #1}\else
  \providecommand{\doi}{doi: \begingroup \urlstyle{rm}\Url}\fi

\bibitem[Blodgett et~al.(2020)Blodgett, Barocas, Daum'e, and
  Wallach]{Blodgett2020LanguageIsPower}
Blodgett, S.~L., Barocas, S., Daum'e, H., and Wallach, H.~M.
\newblock Language (technology) is power: A critical survey of “bias” in
  {NLP}.
\newblock In \emph{Annual Meeting of the Association for Computational
  Linguistics}, 2020.

\bibitem[Brown et~al.(2020)Brown, Mann, Ryder, Subbiah, Kaplan, Dhariwal,
  Neelakantan, Shyam, Sastry, Askell, Agarwal, Herbert-Voss, Krueger, Henighan,
  Child, Ramesh, Ziegler, Wu, Winter, Hesse, Chen, Sigler, Litwin, Gray, Chess,
  Clark, Berner, McCandlish, Radford, Sutskever, and
  Amodei]{Brown2020LanguageModelsAreFewShotLearners}
Brown, T.~B., Mann, B., Ryder, N., Subbiah, M., Kaplan, J., Dhariwal, P.,
  Neelakantan, A., Shyam, P., Sastry, G., Askell, A., Agarwal, S.,
  Herbert-Voss, A., Krueger, G., Henighan, T.~J., Child, R., Ramesh, A.,
  Ziegler, D.~M., Wu, J., Winter, C., Hesse, C., Chen, M., Sigler, E., Litwin,
  M., Gray, S., Chess, B., Clark, J., Berner, C., McCandlish, S., Radford, A.,
  Sutskever, I., and Amodei, D.
\newblock Language models are few-shot learners.
\newblock \emph{ArXiv}, abs/2005.14165, 2020.

\bibitem[Bryant et~al.(2022)Bryant, Yuan, Qorib, Cao, Ng, and
  Briscoe]{bryant2022grammatical_survey}
Bryant, C., Yuan, Z., Qorib, M.~R., Cao, H., Ng, H.~T., and Briscoe, T.
\newblock Grammatical error correction: A survey of the state of the art.
\newblock \emph{arXiv preprint arXiv:2211.05166}, 2022.

\bibitem[Fang et~al.(2023)Fang, Yang, Lan, Wong, Hu, Chao, and
  Zhang]{fang2023chatgpt_highly_fluent_gec}
Fang, T., Yang, S., Lan, K., Wong, D.~F., Hu, J., Chao, L.~S., and Zhang, Y.
\newblock Is {ChatGPT} a highly fluent grammatical error correction system? a
  comprehensive evaluation, 2023.

\bibitem[Garrido-Mu{\~n}oz et~al.(2021)Garrido-Mu{\~n}oz, Montejo-R{\'a}ez,
  Mart{\'i}nez-Santiago, and
  Ure{\~n}a-L{\'o}pez]{GarridoMuoz2021ASO_Survey_Bias}
Garrido-Mu{\~n}oz, I., Montejo-R{\'a}ez, A., Mart{\'i}nez-Santiago, F., and
  Ure{\~n}a-L{\'o}pez, L.~A.
\newblock A survey on bias in deep {NLP}.
\newblock \emph{Applied Sciences}, 11:\penalty0 3184, 2021.

\bibitem[Greshake et~al.(2023)Greshake, Abdelnabi, Mishra, Endres, Holz, and
  Fritz]{Greshake2023MoreTY}
Greshake, K., Abdelnabi, S., Mishra, S., Endres, C., Holz, T., and Fritz, M.
\newblock More than you've asked for: A comprehensive analysis of novel prompt
  injection threats to application-integrated large language models.
\newblock \emph{ArXiv}, abs/2302.12173, 2023.

\bibitem[Kasneci et~al.(2023)Kasneci, Se{\ss}ler, K{\"u}chemann, Bannert,
  Dementieva, Fischer, Gasser, Groh, G{\"u}nnemann, H{\"u}llermeier,
  et~al.]{kasneci2023chatgpt}
Kasneci, E., Se{\ss}ler, K., K{\"u}chemann, S., Bannert, M., Dementieva, D.,
  Fischer, F., Gasser, U., Groh, G., G{\"u}nnemann, S., H{\"u}llermeier, E.,
  et~al.
\newblock {ChatGPT} for good? on opportunities and challenges of large language
  models for education.
\newblock \emph{Learning and Individual Differences}, 103:\penalty0 102274,
  2023.

\bibitem[Lai et~al.(2023)Lai, Ngo, Veyseh, Man, Dernoncourt, Bui, and
  Nguyen]{lai2023chatgpt_beyond_english}
Lai, V.~D., Ngo, N.~T., Veyseh, A. P.~B., Man, H., Dernoncourt, F., Bui, T.,
  and Nguyen, T.~H.
\newblock {ChatGPT} beyond english: Towards a comprehensive evaluation of large
  language models in multilingual learning, 2023.

\bibitem[Nadeem et~al.(2020)Nadeem, Bethke, and Reddy]{Nadeem2020StereoSetMS}
Nadeem, M., Bethke, A., and Reddy, S.
\newblock Stereoset: Measuring stereotypical bias in pretrained language
  models.
\newblock In \emph{Annual Meeting of the Association for Computational
  Linguistics}, 2020.

\bibitem[Ng et~al.(2014)Ng, Wu, Briscoe, Hadiwinoto, Susanto, and
  Bryant]{ng2014conll}
Ng, H.~T., Wu, S.~M., Briscoe, T., Hadiwinoto, C., Susanto, R.~H., and Bryant,
  C.
\newblock The {CoNLL}-2014 shared task on grammatical error correction.
\newblock In \emph{Proceedings of the Eighteenth Conference on Computational
  Natural Language Learning: Shared Task}, pp.\  1--14, 2014.

\bibitem[Nunes et~al.(2023)Nunes, Primi, Pires, Lotufo, and
  Nogueira]{nunes2023evaluating_enem}
Nunes, D., Primi, R., Pires, R., Lotufo, R., and Nogueira, R.
\newblock Evaluating {GPT-3.5} and {GPT-4} models on brazilian university
  admission exams.
\newblock \emph{arXiv preprint arXiv:2303.17003}, 2023.

\bibitem[OpenAI(2023)]{openai2023gpt4}
OpenAI.
\newblock {GPT-4} technical report, 2023.

\bibitem[Perez \& Ribeiro(2022)Perez and Ribeiro]{perez2022ignore}
Perez, F. and Ribeiro, I.
\newblock Ignore previous prompt: Attack techniques for language models, 2022.

\bibitem[Wu et~al.(2023)Wu, Wang, Wan, Jiao, and
  Lyu]{wu2023chatgpt_or_grammarly}
Wu, H., Wang, W., Wan, Y., Jiao, W., and Lyu, M.
\newblock {ChatGPT} or grammarly? evaluating {ChatGPT} on grammatical error
  correction benchmark.
\newblock \emph{arXiv preprint arXiv:2303.13648}, 2023.

\end{thebibliography}
\bibliographystyle{icml2023}

\newpage
\appendix
\onecolumn

\setcounter{table}{0}
\renewcommand{\thetable}{\Alph{section}\arabic{table}}

\section{Appendix -- Example Tables}
\label{sec:example_tables}

\begin{table}[H]
\caption{Examples of TP (True Positive), TN (True Negative), FP (False Positive), FN (False Negative) results.}
\label{table:tp_tn_fp_fn}
\vskip 0.15in
\begin{center}
\begin{small}
\begin{sc}
\begin{tabular}{p{0.31\linewidth}p{0.69\textwidth}}
\textbf{Input phrase} & \textbf{Examples of results} \\ \toprule
\multirow{3}{\linewidth}{\begin{tt}Só encontrei ingressos para a última seção do filme.\end{tt} [incorrect]}
& [TP] \begin{tt}Só encontrei ingressos para a última sessão do filme.\end{tt} \\
 & [FN] \begin{tt}Só encontrei ingressos para a última seção do filme.\end{tt} \\
 & [FN] \begin{tt}Só encontrei ingressos para a última seção daquele filme.\end{tt} \\ \midrule
\multirow{3}{\linewidth}{\begin{tt}Só encontrei ingressos para a última sessão do filme.\end{tt} [correct]}
& [TN] \begin{tt}Só encontrei ingressos para a última sessão do filme.\end{tt} \\
 & [FP] \begin{tt}Só encontrei ingressos para a última seção do filme.\end{tt} \\
 & [FP] \begin{tt}Só encontrei ingressos para a última sessão daquele filme.\end{tt} \\ \midrule
\multirow{3}{\linewidth}{\begin{tt}Mais será possível que ele tenha ido tão longe?\end{tt} [incorrect]}
& [TP] \begin{tt}Mas será possível que ele tenha ido tão longe?\end{tt} \\
 & [FN] \begin{tt}Mais será possível que ele tenha ido tão longe?\end{tt} \\
 & [FN] \begin{tt}Mas será possível que ela tenha ido tão longe?\end{tt} \\ \midrule
\multirow{3}{\linewidth}{\begin{tt}Mas será possível que ele tenha ido tão longe?\end{tt} [correct]}
& [TN] \begin{tt}Mas será possível que ele tenha ido tão longe?\end{tt} \\
 & [FN] \begin{tt}Mais será possível que ele tenha ido tão longe?\end{tt} \\
 & [FN] \begin{tt}Mas será possível que ela tenha ido tão longe?\end{tt} \\ 
 \bottomrule
\end{tabular}
\end{sc}
\end{small}
\end{center}
\vskip -0.1in
\end{table}

\begin{table}[H]
\caption{Examples of identified behaviors in our qualitative analysis.}
\label{table:behaviors}
\vskip 0.15in
\begin{center}
\begin{small}
\begin{sc}
\begin{tabular}{p{0.15\linewidth}p{0.425\linewidth}p{0.425\linewidth}}
\toprule
\textbf{Behavior} & \textbf{Correct Phrase} & \textbf{LLM Result} \\ \midrule
\multirow{3}{\linewidth}{Overcorrection}
& \begin{tt}Quantas vezes vou precisar repetir até você me entender?\end{tt}
& \begin{tt}Quantas vezes vou precisar repetir até você me entender \end{tt} {[}GPT-4{]} \\
& \begin{tt}As obras daquele diretor são impactantes, são filmes que nos fazem refletir de verdade.\end{tt}
& \begin{tt}As obras daquele diretor são impactantes; são filmes que nos fazem refletir de verdade.\end{tt} {[}GPT-3.5{]} \\
& \begin{tt}Podemos conversar, a fim de compreender melhor a questão.\end{tt}
& \begin{tt}Podemos conversar a fim de compreender melhor a questão.\end{tt} {[}GPT-3.5{]} \\ \midrule
\multirow{3}{\linewidth}{Omission}
& \begin{tt}Vou buscar meu relógio no conserto.\end{tt}
& \begin{tt}Vou buscar meu relógio no concerto.\end{tt} {[}GPT-3.5{]} \\
& \begin{tt}Quanto ao recebimento do envelope, confirmei-o na mesma hora.\end{tt}
& \begin{tt}Quanto ao recebimento do envelope, confirmei-lhe na mesma hora.\end{tt} {[}GPT-4{]} \\
& \begin{tt}Vamos construir um bom canal de comunicação, ao invés de continuar com essa dificuldade em estabelecer um diálogo.\end{tt}
& \begin{tt}Vamos construir um bom canal de comunicação, em vez de continuar com essa dificuldade em estabelecer um diálogo.\end{tt} {[}GPT-4{]}
\\ \midrule
\multirow{3}{\linewidth}{Grammatical Miscorrection}
& \begin{tt}Apesar de ela ter deixado todos os direcionamentos, não foi possível realizar a apresentação em sua ausência.\end{tt}
& \begin{tt}Apesar de ela ter deixado todas as orientações, não foi possível realizar a apresentação em sua ausência.\end{tt} {[}GPT-3.5{]} \\
& \begin{tt}O senhor encheu o balde de água.\end{tt}
& \begin{tt}O senhor encheu o balde com água.\end{tt} {[}GPT-3.5{]} \\
& \begin{tt}A astrologia classifica como Mercúrio retrógrado o período pelo qual estamos passando.\end{tt}
& \begin{tt}A astrologia classifica o período pelo qual estamos passando como Mercúrio retrógrado.\end{tt} {[}GPT-3.5{]} \\ \midrule
\multirow{3}{\linewidth}{Ungrammatical Miscorrection}
& \begin{tt}Vou viajar nesse próximo final de semana.\end{tt}
& \begin{tt}Vou viajar neste próximo final de semana.\end{tt} {[}GPT-3.5{]}{[}GPT-4{]} \\
& \begin{tt}Buscaram, então, a melhor decisão para ambos os lados.\end{tt}
& \begin{tt}Buscaram então a melhor decisão para ambos os lados.\end{tt} {[}GPT-3.5{]} \\
& \begin{tt}As duas meninas, que foram tão amigas no passado, hoje já não têm mais contato.\end{tt}
& \begin{tt}As duas meninas que foram tão amigas no passado hoje já não têm mais contato.\end{tt} {[}GPT-3.5{]} \\ \bottomrule
\end{tabular}
\end{sc}
\end{small}
\end{center}
\vskip -0.1in
\end{table}

\section{Appendix -- More Educational Applications}
\label{sec:appendix_more_educational_applications}

In addition to their potential for being a powerful GEC tool for Brazilian Portuguese, LLMs hold promise for improving educational outcomes in Brazilian schools and universities. We list some of these ideas:

\begin{description}[wide=0\parindent]
    \item[Better GEC Tools] LLMs can not only correct ungrammatical phrases but also guide students in understanding why the phrases are ungrammatical and explaining how to fix them. See Table~\ref{table:cot_educational_example} for an example.
    \item[Hyper-personalized tutoring] LLMs can facilitate hyper-personalized tutoring by adapting to each student's unique learning style, progress, and needs, providing tailored feedback and guidance to enhance the learning experience.
    \item[Learning disabilities] LLMs can be used to optimize tools developed to assist students with learning disabilities, creating contents that would give them proper conditions to comprehend subjects, follow instructions and answer questions. 
    \item[Classroom augmentation] LLMs can be integrated into various tools and applications to support classroom teachers, learners, and test developers. For instance, they can be used to generate realistic and diverse practice questions or exercises and to assist in creating engaging and culturally-relevant educational content.
    \item[Grading assistant] LLMs can assist in grading and analyzing students' written responses and provide insights into their understanding and misconceptions.
\end{description}
 
The potential of LLMs extends beyond English language applications, offering numerous possibilities for revolutionizing education in Brazilian Portuguese and other languages.

\begin{table*}[h]
\caption{An example of an application of LLMs for GEC that goes beyond error correction capabilities. Using chain-of-thought, we ask the LLM to 1) check whether the phrase has grammatical errors, 2) correct the phrase in case of grammatical errors, and 3) explain why the phrase is wrong in case of grammatical errors.}
\label{table:cot_educational_example}
\vskip 0.15in
\begin{center}
\begin{small}
\begin{sc}
\begin{tabular}{p{\linewidth}}
\toprule
\textbf{Prompt} \\
\midrule
\begin{tt}You are an agent capable of correcting phrases to grammatically correct Brazilian Portuguese.\end{tt}\\ \\
\begin{tt}Think step by step in the following format:\end{tt}\\
\begin{tt}1. Does the phrase have any grammatical errors?\end{tt}\\
\begin{tt}2. In case of grammatical errors, correct the phrase.\end{tt}\\
\begin{tt}3. Explain why the phrase is wrong.\end{tt} \\ \vspace{0.15cm} \\
\midrule
Phrase: \textit{"João sempre tem um caloroso comprimento para oferecer quando nos encontramos por acaso."} \\ \midrule
\begin{tt}1. Yes, the phrase has a grammatical error.\end{tt}\\
\begin{tt}2. The correct version would be: "João sempre tem um caloroso cumprimento para oferecer quando nos encontramos por acaso."\end{tt}\\
\begin{tt}3. The error in the original phrase was the misspelling of "cumprimento" as "comprimento". "Cumprimento" means "greeting" or "salute", while "comprimento" means "length". Therefore, the original phrase didn't make sense in the context of greeting someone.\end{tt} \\ \vspace{0.15cm} \\
\midrule
Phrase: \textit{"A esposa e o marido tem muito dinheiro."} \\ 
\midrule
\begin{tt}1. Yes, the phrase has a grammatical error.\end{tt}\\
\begin{tt}2. The correct version would be: "A esposa e o marido têm muito dinheiro."\end{tt}\\
\begin{tt}3. The verb "ter" (to have) is conjugated in the third person plural ("têm") to agree with the subjects "a esposa e o marido" (the wife and the husband), which are also in the third person plural. In the original sentence, the verb "tem" is used in the third person singular form, which does not agree with the subjects of the sentence.\end{tt} \\
\bottomrule
\end{tabular}
\end{sc}
\end{small}
\end{center}
\vskip -0.1in
\end{table*}

\end{document}